\documentclass{article}

\usepackage[utf8]{inputenc} 
\usepackage[T1]{fontenc}    
\usepackage{hyperref}       
\usepackage{url}            
\usepackage{booktabs}       
\usepackage{amsfonts}       
\usepackage{nicefrac}       
\usepackage{xfrac}          
\usepackage{microtype}      
\usepackage{wrapfig}

\usepackage{times}
\usepackage{latexsym}
\usepackage{hyperref}
\usepackage{pgfplotstable}
\usepackage{pgfplots}
\usepackage{collcell}
\usepackage{xstring}
\usepackage{tabularx}
\usepackage{graphicx}
\usepackage{tabu}
\usepackage{environ}
\usepackage{fp}
\usepackage{upgreek}
\usepackage{amsmath}
\usepackage{amsthm}
\usepackage{enumitem}
\usepackage{bm}
\usepackage{subfiles}
\usepackage{array,multirow}
\usepackage{caption}
\usepackage{makecell}
\usepackage{tikz, color}
\usetikzlibrary{matrix}
\usetikzlibrary{calc}
\usetikzlibrary{positioning}
\usetikzlibrary{graphs}
\usetikzlibrary{shapes.geometric}

\usepackage{xcolor,colortbl}
\definecolor{matlab-blue}{rgb}{0,0.4470,0.7410}
\definecolor{matlab-orange}{rgb}{0.8500,0.3250,0.0980}
\definecolor{matlab-yellow}{rgb}{0.9290,0.6940,0.1250}

\definecolor{matlab-green}{rgb}{0.4660,0.6740,0.1880}

\definecolor{matlab-red}{rgb}{0.6350,0.0780,0.1840}

\newcolumntype{A}{>{\centering\arraybackslash}m{1cm}}
\newcolumntype{B}{>{\centering\arraybackslash}m{2cm}}
\newcolumntype{C}{>{\centering\arraybackslash}m{1.5cm}}
\newcolumntype{D}{>{\centering\arraybackslash}m{1.2cm}}
\newcolumntype{E}{>{\centering\arraybackslash}m{3cm}}
\newcolumntype{F}{>{\arraybackslash}m{2cm}}
\newcolumntype{G}{>{\centering\arraybackslash}m{2.5cm}}
\newcolumntype{J}{>{\centering\arraybackslash}m{0.8cm}}

\usepackage[accepted]{include/icml2021}
\icmltitlerunning{Learning De-identified Representations of Prosody from Raw Audio}

\pgfplotsset{compat=1.17}
\begin{document}

\twocolumn[
\icmltitle{Learning De-identified Representations of Prosody from Raw Audio}



\icmlsetsymbol{equal}{*}

\begin{icmlauthorlist}
\icmlauthor{Jack Weston}{to}
\icmlauthor{Rapha\"{e}l Lenain}{to}
\icmlauthor{Udeepa Meepegama}{to}
\icmlauthor{Emil Fristed}{to}
\end{icmlauthorlist}

\icmlaffiliation{to}{Novoic, London, UK}

\icmlcorrespondingauthor{Jack Weston}{jack@novoic.com}

\icmlkeywords{Machine Learning, ICML}

\vskip 0.3in
]



\printAffiliationsAndNotice{} 

\begin{abstract}
We propose a method for learning de-identified prosody representations from raw audio using a contrastive self-supervised signal. Whereas prior work has relied on conditioning models on bottlenecks, we introduce a set of inductive biases that exploit the natural structure of prosody to minimize timbral information and decouple prosody from speaker representations. Despite aggressive downsampling of the input and having no access to linguistic information, our model performs comparably to state-of-the-art speech representations on DAMMP, a new benchmark we introduce for spoken language understanding. We use minimum description length probing to show that our representations have selectively learned the subcomponents of non-timbral prosody, and that the product quantizer naturally disentangles them without using bottlenecks. We derive an information-theoretic definition of speech de-identifiability and use it to demonstrate that our prosody representations
 are less identifiable than other speech representations.
\end{abstract}
\section{Introduction}\label{sec:introduction}

To produce and understand spoken language, humans encode and decode audio information at multiple timescales. Phonetic information is used to decode linguistic content, which carries part of the meaning in speech. Another source of meaning, prosody, can be decoded through non-phonetic acoustic patterns; for example to identify who's speaking, which primarily relies on the timbre subcomponent of prosody \cite{skerry2018towards,qian2020unsupervised}. Artificial systems mimicking these human processes must solve the similar problems of representing phonetic information (to obtain a linguistic representation) and representing prosodic information \cite{baevski2020wav2vec,kenter2019chive}.

The phonetic problem has automatic speech recognition (ASR) as its obvious use-case. In recent years, speech representation learning has been increasingly dominated by self-supervised frameworks using contrastive losses. These include contrastive predictive coding (CPC) \cite{oord2018representation}, wav2vec \cite{schneider2019wav2vec}, vq-wav2vec \cite{baevski2019vq} and wav2vec 2.0 \cite{baevski2020wav2vec}, which learn representations directly from raw audio. These models are generally frame- or phone-based and use fine-timescale, fixed-length audio frames as input. This promotes encoding of high-frequency phonetic information, crucial for transcription, but makes them less incentivized to capture patterns occurring on longer timescales. Other approaches have used triplet loss and temporal proximity as a training signal to learn ``semantic'' \cite{jansen2018unsupervised} or ``non-semantic'' \cite{shor2020towards} representations from spectrograms.

The prosodic problem has been less studied. Its primary use-case has been building more expressive text-to-speech (TTS) systems. Prior approaches to learning representations of prosody have relied on subtractive definitions such as \cite{skerry2018towards}: ``Prosody is the variation in speech signals that remains after accounting for variation due to phonetics, speaker identity, and channel effects (i.e. the recording environment)''. These approaches typically use autoencoders conditioned on lexical information and speaker identity \cite{skerry2018towards, wang2018style, battenberg2019effective, zhang2019learning}. This encourages the remaining information to be contained in a bottleneck that encodes prosody. Other work has used a triple bottleneck to further decompose prosody in its subcomponents \cite{qian2020unsupervised}. For \emph{non-timbral prosody} (i.e. what remains after removing speaker characteristics), these subcomponents are pitch, rhythm and tempo, acoustically reflected in the fundamental frequency ($F_0$, the pitch contour), intensity or energy ($c_0$), and the speech rate respectively. WaveNet makes explicit use of $F_0$, $c_0$ and phone durations to synthesize speech \cite{oord2016wavenet}. CHiVE represents prosody using $F_0$, $c_0$ and phoneme durations as features in a conditional variational autoencoder \cite{kenter2019chive}. These models share a set of characteristics that motivate our work: they are subtractive and rely on conditioning models on bottlenecked information. Having been developed in the context of TTS, they have inductive biases that are particularly suited to learning primarily phonetic representations, rather than prosodic representations, a weakness highlighted by \citet{oord2016wavenet}.

The contributions of this paper are as follows:
\begin{itemize}
    \item Introducing and characterizing VQP, a self-supervised contrastive model that learns to selectively represent non-timbral prosody from raw audio without using bottlenecks.
    \item Adapting probes from the natural language processing (NLP) literature to demonstrate that VQP representations selectively encode the subcomponents of non-timbral prosody.
    \item Demonstrating that product quantization can be used for disentanglement of audio representations without using bottlenecks.
    \item Introducing an information-theoretic definition of de-identification using prequential probes on a speaker verification task.
    \item Benchmarking a number of state-of-the-art, self-supervised audio representation learning models on a set of tasks for spoken language understanding, as well as quantifying their de-identifiability.
\end{itemize}


\section{Approach}\label{sec:approach}
\subsection{Data Preprocessing}
Before data is passed to the model, it undergoes a series of preprocessing steps. The raw audio is first resampled to $16\,$kHz, then pitch-shifted on a per-example basis such that the median pitch of the voice segments is the same value for the whole dataset. To perform this pitch-shifting, we ran an autocorrelation-based pitch-tracking method (via Praat \cite{boersma2006praat}), calculated the median pitch of the voiced segments in a given sample, then shifted the pitch such that the median is $150\,$Hz for the sample.  This is primarily to address the distributional difference between sexes in fundamental frequency. We find that this makes the resulting representations less identifiable and improves training robustness. The waveform is aggressively downsampled to $500\,$Hz, retaining the frequency range of the typical fundamental frequency for human speech, but discarding the formants that characterise the speaker's voice. The highest typical $F_0$ for humans is $\sim250\,$Hz \cite{takefuta1972statistical} and by applying Nyquist's theorem \cite{nyquist1928certain} we choose a sampling rate of $500\,$Hz. The waveform is normalized to zero mean and unit variance, then sliced into variable-length audio-words using word-level timestamps obtained via ASR or forced alignment. This is based on the intuition that meaningful prosody states are naturally discretized on a per-word basis. By contrast, wav2vec, vq-wav2vec, wav2vec 2.0, TRILL, Tacotron and CHiVE all use fixed-length audio input \cite{schneider2019wav2vec, baevski2019vq, baevski2020wav2vec, shor2020towards, wang2017tacotron, kenter2019chive}. In \citet{stehwien2017prosodic}, word-level timestamps are used with spectrogram inputs to a CNN on a classification task to recognise prosodic events. Prior models, developed for ASR or TTS use shorter timescale audio inputs \cite{baevski2019vq, skerry2018towards}. We include up to $2$ seconds of leading silence before each audio word since pause information is important for parts of prosody such as the rhythm and tempo of speech.

\subsection{Model}

\begin{figure*}[ht]
\vskip 0.2in
\begin{center}
\centerline{\includegraphics[width=0.95\textwidth]{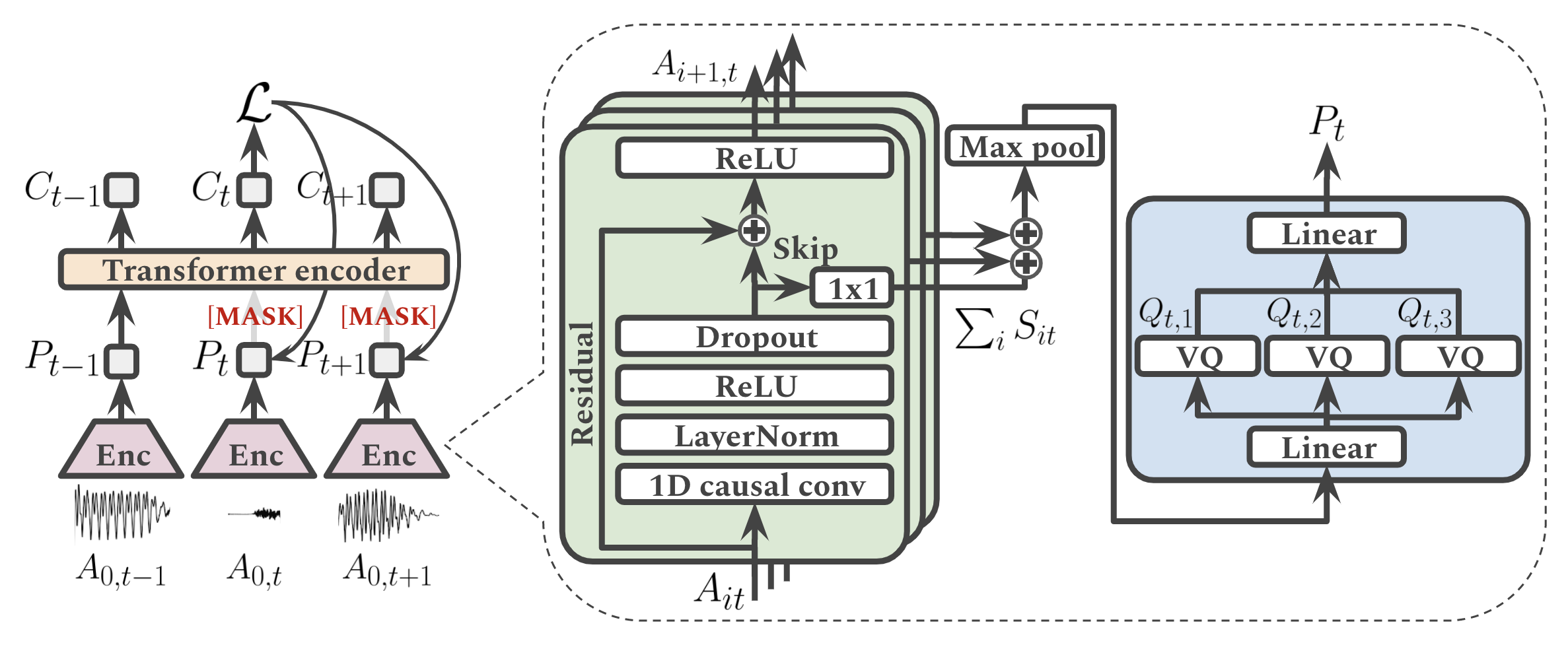}}
\caption{The model architecture, including an overview of the entire model (left) and a focus on the prosody encoder (right).}
\label{fig:architecture}
\end{center}
\vskip -0.2in
\end{figure*}

Our model comprises two parts: a prosody encoder and a Transformer encoder (see Figure \ref{fig:architecture}). The prosody encoder $f: \mathcal{A} \mapsto \mathcal{P}$ maps variable-length raw audio $A_{0,t}$, corresponding to a single audio-word, to a fixed-length quantized vector $P_{t}$. The sequence of latent prosody representations $P_{t}$ is fed to a Transformer $f: \mathcal{P} \mapsto \mathcal{C}$ to produce contextualized prosody representations $C_{t}$ that can capture information from the entire sequence, unlike the audio-word-level $P_{t}$ representations. Our core hypothesis is that prosody has predictable temporal patterns, occurring at frequencies lower than $250\,$Hz that can be learned directly from the acoustic signal. We use a contrastive, self-supervised signal, similar to \citet{baevski2020wav2vec}, where only raw audio is used as the input and target. Unlike subtractive approaches to representing prosody \cite{skerry2018towards, kenter2019chive}, our model does not rely on lexical inputs. Instead, it only has access to the downsampled raw audio signal and word-level timestamps.

\textbf{Temporal convolutional network:} The first module of the prosody encoder is a temporal convolutional network (TCN) \cite{oord2016wavenet, bai2018empirical}, comprising a stack of causal dilated 1D convolutions with residual connections, which we adapt with skip connections \cite{he2016deep}. The strides, number of layers and kernel sizes are chosen such that the receptive field of the TCN spans the maximum sequence length of one audio word. Inspired by the WaveNet architecture \cite{oord2016wavenet}, we use summed skip-connections as the TCN output rather than the output of the final layer  to allow the network to more easily capture features with different time-resolutions. To reduce across the temporal (frame) dimension, we max-pool the skip matrix, which we empirically found led to more robust training than mean pooling or selecting the final non-padded element in the skip matrix for each element in the batch. We exploit the exponentially increasing receptive field of our TCN to capture the longer-range dependencies that encode prosodic information. By contrast, other approaches to encoding prosody have relied on conditional autoencoders \cite{skerry2018towards, kenter2019chive, zhang2019learning} or CNNs \cite{stehwien2017prosodic}. TTS-motivated models encoding phonetic information have used linguistic features such as $F_0$, duration and $c_0$ \cite{kenter2019chive, oord2016wavenet}. ASR-motivated models have used raw audio passed into a TCN \cite{baevski2020wav2vec} or a vanilla CNN \cite{baevski2019vq}.

\textbf{Product quantizer:} The max-pooled output of the TCN is passed to a product quantizer, whose constituent vector quantizers are inspired by VQ-VAE-2 \citet{razavi2019generating} but adapted for product quantization. The product quantizer itself is similar to wav2vec 2.0, wherein the input data undergoes an affine transformation before having its features sliced into $M$ equal parts, all of which are passed to a vector quantizer. Following quantization, the output vectors are concatenated and undergo a final affine transformation. Following \cite{razavi2019generating}, each constituent vector quantizer learns a nonlinear mapping from its input space $S$ to a vector $E(\bm{s})$, which is replaced with the nearest prototype vector in the codebook $\bm{e}_k, k \in 1, \ldots, K$:
\begin{equation}
\mathrm{Quantize}\left(E\left(\bm{s}\right)\right) = \bm{e}_k,
\end{equation}
where $k={\operatorname{arg}\,\operatorname{max}}_j \left\lVert E\left(\bm{s}\right) - \bm{e}_j \right\rVert$. This mapping is learnt via backpropagation using the straight-through gradient estimator \cite{bengio2013estimating}. Using multiple vector quantizers is not equivalent to using one with a larger capacity; the inclusion of affine transformations before and after the vector quantization gives the network some capacity to map the input data into a more convenient basis before slicing. We will explore whether product quantization can be used to disentangle the space of representations and add explainability to representation learning models. The use of VQ-VAE-style quantization over the Gumbel softmax \cite{jang2016categorical} approach used in wav2vec 2.0 was an empirical decision to improve training robustness. This is consistent with the claim in the original VQ-VAE paper \cite{oord2017neural} that their method experiences less gradient variance during training. We deliberately restrict the number of quantized states in our codebook while learning the vector-quantized representations to encourage representations to be parsimonious and avoid ``hiding'' nuisance covariates, which may include speaker-identifiable information, in small details. 

\textbf{Transformer encoder:} The product-quantized vector sequence is fed to a standard Transformer  encoder architecture \cite{vaswani2017attention}. We use fixed sine/cosine positional embeddings to allow the encoder to exploit position information. In wav2vec \cite{schneider2019wav2vec}, representations of audio data are learned by solving a self-supervised context-prediction task with the same loss function as word2vec \cite{mikolov2013efficient}. One aim of contextualizing prosody representations is to make representations with weaker cross-temporal interactions, which may facilitate audio-linguistic representation learning in future work. Context-aware representations of time-series often make better predictions \cite{peters2018deep, devlin2018bert, oord2018representation} and we hypothesise that contextualization makes stronger prosodic representations.

\subsection{Probing and De-identification}
For explainability, we aim to measure how well a feature is represented in a given representation. We use the prequential (or online) approach to minimum description length (MDL) probing to quantify the regularity between representations and labels \cite{rissanen1978modeling, voita2020information}. Formally, MDL measures the number of bits required to transmit the labels given the representations. If a feature is easily extractable from a given representation, a model trained to detect said feature will converge quickly, resulting in a small MDL. Computing the MDL using the prequential approach requires sequential training and evaluation. We partition the train set, $D=\{(x_i, y_i)\}^{n}_{i=1}$, into timesteps, $1=t_0<t_1<\hdots<t_S=n$, and train our probe, $p_{\theta}(y|x)$, such that at timestep $t_i$ the train set is $\{(x_j, y_j)\}^{t_i}_{j=1}$ and we evaluate on set $\{x_j, y_j\}^{t_{i+1}}_{j=t_i+1}$, calculating the codelength as per \citet{voita2020information}.

We further adapt this method to derive an information-theoretic definition of speech identifiability. Following the literature \cite{tomashenko2020introducing}, we consider a number of binary speaker verification trials but, instead of using equal error rate or log-likelihood-based metrics, we define the \emph{de-identification ratio} of a set of trial representations $\{s_i\}$ with respect to enrolment representations $\{r_i\}$ as the inverse of the compression ratio of  the theoretical minimum description length to transmit the data using a prequential approach:
\begin{align}
    &\mathcal{D}\left(s, r, M(\theta), D\right) = \nonumber \\
    & \frac{t_1}{n} - \frac{1}{n}\sum_{i=1}^{S-1} \log_2 p_{\theta_i}\left(y_{t_i+1:t_{i+1}} | r_{t_i+1:t_{i+1}}, s_{t_i+1:t_{i+1}}\right).
\end{align}
A full derivation is given in the supplementary materials. The rationale is that a shorter MDL means that the verification task is easier given the two representations. This improves upon prior work, which assumes a fixed model  (usually a probabilistic LDA \cite{tomashenko2020introducing, han2020voice}), by taking into account the effort required to perform verification as well as performance on the task. Real attackers could have access to sophisticated models and arbitrary computational resources to compare speech representations, motivating this approach. Prior work performs verification on pairs of i-vectors \cite{tomashenko2020introducing}; we likewise consider pairs of the same representation, but note that cross-representation comparisons ought to be included in more comprehensive studies, including raw audio and spectrogram as inputs. For simplicity, we mean-pool sequential representations over time but note that this could underestimate the identifiability of the sequence as a whole due to lost information.
\section{Experimental Setup}\label{sec:experiemental_setup}

\subsection{The Colossal Audio-Linguistic Corpus (CALC)}\label{subsec:calc}

Models using self-supervised pretraining consistently demonstrate the importance of large datasets \cite{devlin2018bert, raffel2019exploring, brown2020language}. AudioSet \cite{gemmeke2017audio}, for instance, is a large dataset for general-purpose audio machine learning, a significant subset of which has speech tags. For pretraining our models, we construct the Colossal Audio-Linguistic Corpus (CALC), a large word-aligned audio-linguistic dataset of natural speech with matching audio and text modalities. CALC is composed of five datasets wrangled into a common format,  chosen based on their size, prior use in the literature, and whether they contain natural speech as opposed to read speech (Table \ref{tab:calc}). Except for the AMI dataset, which already has word-level alignment \cite{carletta2005ami}, we performed alignment using the Montreal Forced aligner \cite{mcauliffe2017montreal}. The full CALC dataset contains $\sim7.5$ million words of natural speech.

\begin{table*}[t]
    \centering
    \begin{tabular}{A A B G G G B B}
    \toprule
        CALC & DAMMP & Dataset & Target & Description & Size & Ref. \\
        \midrule
        - & $\surd$ & DAIC-WOZ & Depression diagnoses & Interviews by a virtual interviewer & ${\scriptsize \sim}$300 interviews & \cite{gratch2014distress}\\  
        - & $\surd$ & ADReSS  & Alzheimer's disease diagnoses & Picture description tasks & ${\scriptsize \sim}$200 descriptions & \cite{luz2020alzheimer}\\  
        - & $\surd$ & MUStARD   & Sarcasm labels & Acted scenes from TV shows & ${\scriptsize \sim}$6.4k utterances & \cite{castro2019towards}\\  
        $\surd$ & $\surd$ & CMU-MOSEI  & Sentiment labels &  Spoken product reviews & ${\scriptsize \sim}$20k utterances from ${\scriptsize \sim}$2k speakers  & \cite{zadeh2018multimodal}\\  
        $\surd$ & $\surd$ & POM  & Persuasiveness labels & Film reviews & ${\scriptsize \sim}$300 reviews & \cite{park2014computational}\\  
        $\surd$ & - & TED-LIUM 3 &  - & TED talks & ${\scriptsize \sim}$2.4k TED talks & \cite{hernandez2018ted}\\  
        $\surd$ & - & LRS2  & - & Single utterances from BBC TV scenes & ${\scriptsize \sim}$140k utterances & \cite{afouras2018deep}\\     
        $\surd$ & - & AMI  & - & Real and acted meetings & ${\scriptsize \sim}$100 hours of meetings & \cite{carletta2005ami}\\           
    \bottomrule
    \end{tabular}
    \caption{The source datasets for the Colossal Audio-Linguistic Corpus (CALC) and the DAMMP benchmark for spoken language understanding, along with a short description of the nature and size of the data. CMU-MOSEI and POM feature in both datasets due to their size and the existence of relevant targets. The DAMMP benchmark defines a set of supervised classification tasks, the targets of which are shown in the table.}
    \label{tab:calc}
\end{table*}

\subsection{The DAMMP Benchmark for Spoken Language Understanding}\label{subsec:dammp}

To standardize assessment of representations for spoken language understanding, we introduce a new benchmark, DAMMP. The dataset has parallel audio and text modalities of natural speech, so audio-based, text-based and audio-linguistic models can be benchmarked. DAMMP is composed of five datasets (Table \ref{tab:calc}) all with binary classification tasks where prosody is important: DAIC-WOZ \cite{low2020depression}, ADReSS \cite{garcia2020artificial,pompili2020inescid}, MUStARD \cite{bryant2005mustard,woodland2011mustard}, CMU-MOSEI \cite{Liu2018MOSEI,jain2018mosei}, and POM \cite{okada2016pom,siddiquie2015pom}. As with CALC, we performed word-level alignment for all datasets using the Montreal Forced aligner \cite{mcauliffe2017montreal}. DAIC-WOZ, ADReSS, and CMU-MOSEI already had a canonical test set disjoint by speaker, whereas for MUStARD and POM we sampled the datasets to make balanced test sets for the binary variable of interest. For MUStARD, the train/test sets are disjoint by TV show as well as speaker, to make the task harder. DAIC-WOZ, ADReSS and MUStARD already had canonical binary labels to predict. For CMU-MOSEI and POM, we converted Likert-scale ratings (persuasiveness and sentiment respectively) to binary labels, following \citet{park2014computational}.

\subsection{Baselines}
We compare VQP’s performance on DAMMP and on de-identification with four recent audio representation learning models: TRILL \cite{shor2020towards}, wav2vec 2.0 \cite{baevski2020wav2vec}, vq-wav2vec \cite{baevski2019vq} and Mockingjay \cite{liu2020mockingjay}. We chose these baselines as they each bear similarity to VQP in different ways, though we note that we were unable to obtain a pure prosodic baseline. An improvement for future work would be to compare with a baseline from a TTS model, such as \citet{skerry2018towards, kenter2019chive, wang2018style}. Additionally, we compare against x-vectors \cite{snyder2018spoken, SB2021} which are specifically trained to learn identifiable representations through speaker identification pretraining.


\subsection{Training}
The model is trained using a self-supervised contrastive signal, followed by assessing performance on a supervised task. The representations are not fine-tuned on the supervised task to preclude the model from pulling out new, perhaps identifiable, information from the raw audio during supervision. With that caveat, we will still refer to the self-supervised step as `pretraining' for convenience.

We pretrain using a BERT-like masking paradigm, with a contrastive self-supervised signal similar to wav2vec 2.0. The pretraining task is to identify the correct latent prosody representation in the presence of a number of distractors sampled from other masked timesteps. We mask timesteps with a fixed probability and consider a two-part loss function: a contrastive loss and a commitment loss,
\begin{equation}
\mathcal{L} = \mathcal{L_\mathrm{contrast}} + \alpha \mathcal{L_\mathrm{commit}},
\end{equation}
where $\alpha$ is a tunable constant. The contrastive loss for selecting the true latent prosody representation $\bm{q}_t$ amongst a set of $K$ distractors $\widetilde{\bm{q}}_t \in \widetilde{\bm{Q}_t}$, which are uniformly sampled from other masked timesteps of the same sample, is given by:
\begin{equation}
    \mathcal{L_\mathrm{contrast}} = - \log \frac{\exp \left[ \bm{c}_t^T\bm{q}_t / \left( \kappa \left\lVert \bm{c}_t \right\rVert\left\lVert \bm{q}_t \right\rVert \right)\right]}{\sum_{\bm{\widetilde{q}}\sim \bm{\widetilde{Q}}_t}\exp \left[ \bm{c}_t^T\bm{\widetilde{q}} / \left( \kappa \left\lVert \bm{c}_t \right\rVert\left\lVert \bm{\widetilde{q}} \right\rVert \right)\right]},
\end{equation}
where $\kappa$ is a tunable constant. The commitment loss penalizes discrepancies between the quantizer inputs and outputs to encourage robustness. We average the commitment loss over the $N$ constituent vector quantizers in our product quantizer:

\begin{equation}
    \mathcal{L_\mathrm{commit}} = \frac{1}{N} \sum_{i=1}^N\left\lVert \mathrm{sg}\left[\bm{e}_i\right] - E_i\left(\bm{x}\right) \right\rVert^2_2,
\end{equation}
where $\mathrm{sg}\left(\cdot\right)$ is the stop gradient operator and $\bm{x}$ is the training example. In lieu of a codebook loss, we use exponential moving average updates for the codebook as per \citet{oord2017neural}.

For training on downstream tasks, we use a simple two layer feed-forward network (FFN) with hidden size $256$, batch size $256$, ReLU activations, dropout with probability $30\%$ using the Adam optimizer \cite{kingma2014adam} with learning rate $\alpha = 10^{-3}$ and default parameters $\beta_1=0.9, \beta_2=0.99$. We use a final sigmoid nonlinearity and binary cross-entropy loss. The input dimension varies across the different representations. We train these models for 20k steps and use the last model states to report performance on the downstream tasks.

\subsection{Experiments}

We pretrain our models using a proprietary framework built on top of PyTorch \cite{paszke2019pytorch}. We uniformly mask $30\%$ of all prosody tokens. The TCN comprises $9$ layers, each with $30$ filters, a stride of $1$ and a kernel size of $2$. We use exponentially increasing dilations of size $1, 2, 4, 8, 16, 32, 64, 128, 256$ to yield a receptive field size of 512 frames. The $1\times1$ convolution similarly has $30$ filters. The dropout probability is $10\%$.

The product quantizer comprises $3$ vector quantizers each of dimension $10$ with an independent codebook of size $32$, giving a maximum number of states of $32 \times 32 \times 32 \approx 32.8$k per audio-word. We experiment with fewer and more states (see supplementary materials). We choose a decay of $\gamma = 0.99$ for all quantizers and weight the commitment loss by $\alpha = 0.5$. The linear layers have dimensionality $30$.

The Transformer encoder has $12$ layers, $12$ attention heads, inner (FFN) dimension $3,072$, embedding size $768$, ReLU activation and a $10\%$ dropout probability. The positional encoding is implemented as per \citet{vaswani2017attention}. We postulate that prosody temporal interactions are relatively short compared to language and restrict the sequence length to $32$ words. During pretraining, we also require a minimum sequence length of $16$ words. We train using $K=9$ distractors.

We linearly warm up the learning rate from $0$ to a maximum of $1.5 \times 10^{-5}$ at $10$k steps before linearly decaying it to $0$ at the step. The model trains for $250$k steps using the AdamW optimizer \cite{loshchilov2017decoupled}. We use a batch size of $128$ samples and train on a single V100 GPU for 2.3 days.

\section{Results and Discussion}\label{sec:results}

\begin{table*}[ht]
    \centering
    \begin{tabular}{FCCCCC|CC}
    \toprule
         & DAIC-WOZ & MOSEI & MUStARD & POM & ADReSS & DAMMP score & DIR\\
        \midrule
        VQP (ours) &\textbf{0.667} &0.659 &0.488 &0.700 &0.652 & 0.633 & \textbf{1.10} \\
        TRILL &0.619 &0.819 &0.546 &0.684 &\textbf{0.770} & 0.687 & 0.90\\
        wav2vec-2.0 &0.541 &\textbf{0.835} &0.586 &0.722 &0.621 & 0.661 & 0.56 \\
        vq-wav2vec &0.521 &0.831 &\textbf{0.618} &0.792  &0.685 & \textbf{0.687} & 0.84 \\
        Mockingjay &0.463 &0.829 &0.504 &0.785 &0.580 & 0.632 & 0.60 \\
        x-vector &0.491 &0.778 &0.567 &\textbf{0.877} &0.540 & 0.651 & 0.36 \\
        &\\
        \multicolumn{4}{l}{\underline{Ablations}} \tabularnewline
        VQP-PE &0.537 &0.643 &0.489 &0.647 &0.64 & 0.591 & 0.96 \\
        VQP-CSN &0.491 &0.694 &0.568 &0.51 &0.295 & 0.512 & 0.56 \\
        VQP-NPS &0.496 &0.63 &0.551 &0.702 &0.639 & 0.604 & 1.05 \\
        VQP-NQ &0.587 &0.69 &0.564 &0.668 &0.641 & 0.630 & 1.15 \\
        VQP-TQ &0.580 &0.510 &0.523 &0.681 &0.607  & 0.580 & 1.18 \\
        &\\
        \multicolumn{4}{l}{\underline{500\,Hz filter experiments}} \tabularnewline
        TRILL & 0.405 & 0.526 & 0.522 & 0.375 & 0.500  & 0.466 & 1.04 \\  
        wav2vec-2.0 & 0.452 & 0.554 & 0547 & 0.495 & 0.623 & 0.534 & 1.00 \\  
        vq-wav2vec & 0.693 & 0.556 & 0.499 & 0.444 & 0.478  & 0.534 & 1.03 \\  
        Mockingjay & 0.498 & 0.561 & 0.512 & 0.486 & 0.518  & 0.515 & 1.01 \\  
    \bottomrule
    \end{tabular}
    \caption{The results of our work (VQP), baseline representations and ablations on the DAMMP benchmark and the de-identification ratio (DIR). We report AUCs for each constituent classification task and define the DAMMP score as the average. VQP-PE = VQP using the prosody encoder output rather than the Transformer output; VQP-CSN = VQP where the negatives prosodies are cross-sampled, i.e. from other utterances (usually other people); VQP-NPS = VQP trained on data without the median pitch scaling preprocessing step; VQP-NQ = VQP with no product/vector quantization;  VQP-TQ = VQP with only targets quantized (similar to wav2vec-2.0).}
    \label{tab:results}
\end{table*}

\subsection{Benchmarking Performance and De-identifiability}
The results on DAMMP and the de-identifiability task are given in Table \ref{tab:results} for this work (VQP), baselines and ablations. No model uniformly performed the best across all benchmark tasks. VQP performed the best on DAIC-WOZ (AUC = $0.667$), TRILL the best on ADReSS (AUC = $0.770$), wav2vec-2.0 on MOSEI (AUC = $0.835$), vq-wav2vec on MUStARD (AUC = $0.618$), and x-vector on POM (AUC = $0.877$). Averaging the AUCs across all datasets (which we term the DAMMP score), vq-wav2vec performed the best (score = $0.689$) and TRILL comparably (score = $0.687$). VQP had an average score of $0.633$. The observation that different representations performed the best on different tasks highlights the complexity of spoken language understanding, and the scope for improved representations in the area.

VQP had the worst performance compared to other models on MOSEI and MUStARD. For MOSEI, examples are single utterances of $1-20$ words, about an order of magnitude shorter by words than the other tasks. Since VQP produces word-level representations whereas other baselines work at a finer timescale, we hypothesize that the time-averaging did not have enough examples to produce a high-quality pooled representation for VQP. For MUStARD, we hypothesize that the presence of laughter following sarcastic remarks might be used as a training signal for the baselines (which consider all audio), whereas VQP only has access to audio that corresponds to spoken words and up to $2$ seconds of preceding silence, which neglects the laughter which primarily occurs after the final word.

Our ablations show that contextualization improved VQP performance, providing support for our belief that prosody is context-dependent. Pitch scaling had a significant effect on performance, potentially through making the model converge earlier. Removing the product quantizer only led to a small drop in average AUC but, dissecting further, performance dropped substantially on all tasks, except for MUStARD and MOSEI, where it improved. Only quantizing the targets significantly hurt performance, which is notable since it had a positive impact in the original wav2vec-2.0 model \cite{baevski2020wav2vec}. Some of these effects could be reflections of causing the models to converge faster or slower and training models for a fixed number of updates.

VQP outperforms the baselines as the most de-identified, with a codelength of $51.06\,$kbits, de-identification ratio $1.10$ and a probe AUC of $0.73$, compared to the most identifiable baseline, x-vector, that has a codelength of $16.16\,$kbits, a DIR of $0.36$, and an AUC of $0.99$. With the caveat that using performance metrics like AUC directly makes conclusions more strongly model-dependent, we can create a simulation by assuming that we had a speech representation from each of a group of $N$ people and wish to find out from whom a separate target speech representation came. For simplicity, we assume that the model outputs a binary value, the trials are independent and that we must uniquely identify the correct person. Using the classifiers' positive and negative predictive values, we have that $P_\mathrm{id}(N) = \mathrm{PPV} \times \mathrm{NPV}^{N-1}$. For $N=10$ people, VQP would have a probability of correctly identifying the speaker of $1.58\%$, TRILL $5.10\%$, vq-wav2vec $24.8\%$, wav2vec-2.0 $37.7\%$, Mockingjay $44.3\%$ and x-vector $66.1\%$.

Our ablations demonstrate that contextualization helps with de-identification, perhaps by building more abstract representations of prosody that retain less of the identifiable low-level information, such as absolute pitches and tempos. Using cross-sample negatives dramatically hurts de-identifiability, in line with our expectations: if trained on negative examples from other speakers, the representations are encouraged to learn more speaker-specific information. Pitch scaling only had a small effect on de-identifiability, contrary to our expectation. Both ``no quantization'' and    ``quantizing target only'' appeared to improve de-identification. One explanation of this is that the models generally predict less well, which is substantiated by the ablation results on DAMMP, highlighting the trade-off between performance and identifiability. Contrary to our expectations, quantization appears to damage de-identification performance. We hypothesize that this is due to the non-quantized networks having more complex training dynamics, as we observed quantization slowed convergence dramatically.

One ablation that couldn't be done without materially altering the architecture of VQP due to GPU memory constraints, was the importance of the input downsampling. Instead, we applied the $500\,$Hz filter to the DAMMP data before upsampling back to $16\,$kHz, which we then used for the downstream tasks of the baselines (Table \ref{tab:results}). We observed near-random performance on the DAMMP benchmark and high de-identifiability but we believe this was because the preprocessing incurred too great a domain shift for the baselines,  evidenced by the destruction of non-timbral prosodic as well as timbral information, which we probed via an explainability study similar to Section \ref{subsec:explainability}.

\begin{figure}[ht]
\vskip 0.2in
\begin{center}
\hspace*{-0.2in}
\centerline{\includegraphics[width=0.99\columnwidth]{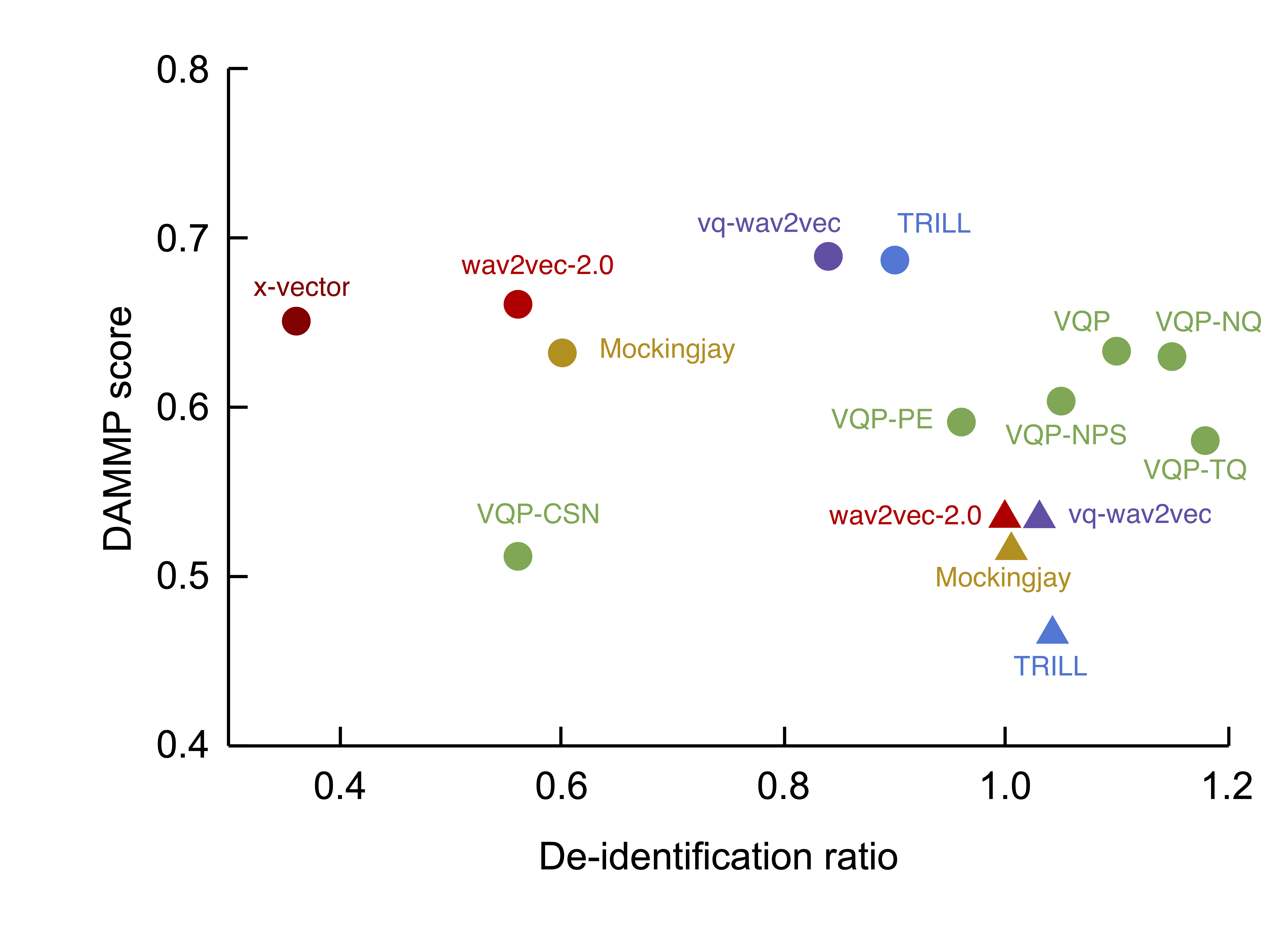}}
\caption{Comparison of VQP, baseline and ablation representations on the DAMMP benchmark and de-identifiability ratio. Triangles represent results obtained using our baselines after downsampling the audio to $500\,$Hz before upsampling back to $16\,$kHz. See Table \ref{tab:results} for a definition of the VQP ablation abbreviations and the DAMMP score.}
\label{fig:comparison}
\end{center}
\vskip -0.2in
\end{figure}

In Figure \ref{fig:comparison}, we investigate the tradeoff between performance and de-identifiability. While wav2vec-2.0 and x-vector perform strongly on the DAMMP benchmark, they are significantly more identifiable than VQP, TRILL and vq-wav2vec. Similarly, while VQP and its variants tend to be much more de-identified, this is traded off with a suppressed performance. The optimal representation for a given context depends jointly on the complexity of the task and the level of privacy required.

\subsection{Using MDL Probes for Explainability}\label{subsec:explainability}

\begin{table*}[t]
    \centering
    \begin{tabular}{lAAAAAAAA|AAAA}
    \toprule
        \multirow{2}{*}{} &  \multicolumn{2}{c}{TRILL} &\multicolumn{2}{c}{wav2vec-2.0} & \multicolumn{2}{c}{vq-wav2vec} & \multicolumn{2}{c}{Mockingjay} & \multicolumn{2}{c}{VQP} \\
        & AUC & MDL & AUC & MDL & AUC & MDL & AUC & MDL & AUC & MDL\\
        \midrule
        \textbf{Pitch} & & & & & & & & \\
        Pitch         &  0.558 &  63.65 &       0.546 &  63.88 &      0.569 &  63.49 &      0.558 &  63.62 & 0.742 & \textbf{55.78}\\
        \midrule
        \textbf{Rhythm} & & & & & & & & \\
        Intensity     &  0.596 &  63.48 &       0.557 &  64.19 &      0.567 &  64.10 &      0.558 &  64.20 & 0.662 & \textbf{60.97}\\
        Num. sylls    &  0.519 &  65.51 &       0.508 &  65.58 &      0.516 &  65.48 &      0.513 &  65.50 & 0.616 & \textbf{63.13}\\
        \midrule
        \textbf{Tempo} & & & & & & & & \\
        Artic. rate   &  0.522 &  \textbf{65.19} &       0.506 &  65.26 &      0.514 &  \textbf{65.19} &      0.510 &  65.29 & 0.537 & \textbf{65.12}\\
        Speech rate   &  0.532 &  \textbf{64.94} &       0.515 &  65.03 &      0.519 &  64.97 &      0.519 &  65.01 & 0.541 & \textbf{64.88}\\
        Syll duration &  0.524 &  \textbf{65.44} &       0.509 &  65.52 &      0.513 &  65.48 &      0.508 &  65.49 & 0.497 & \textbf{65.47}\\
        Word duration &  0.544 &  65.40 &       0.522 &  65.58 &      0.539 &  65.47 &      0.536 &  65.50 & 0.749 & \textbf{54.58}\\
        \midrule
        \textbf{Timbre} & & & & & & & & \\
        Formant f1    &  0.735 &  \textbf{58.03} &       0.668 &  62.73 &      0.696 &  61.26 &      0.629 &  64.07 & 0.574 & 65.58\\
        Formant f2    &  0.743 &  \textbf{57.43} &       0.643 &  63.11 &      0.666 &  62.95 &      0.586 &  64.87 & 0.514 & 65.60\\
        Formant f3    &  0.779 &  \textbf{54.39} &       0.667 &  62.24 &      0.688 &  61.92 &      0.623 &  63.90 & 0.509 & 65.71\\
    \bottomrule
    \end{tabular}
    \caption{The explainability results from probing the VQP representations and the baselines. We probe by performing higher-than-mean/lower-than-mean binary classification on a set of speech features. We report the minimum description length (MDL) for each classification, along with the AUC on the final tranche of data using the prequential approach for comparison. For AUC, a higher value suggests the information is better-represented. For MDL, lower values suggest this. We indicate in bold the minimum MDL values for each speech feature.}
    \label{tab:explainability}
\end{table*}

\begin{table*}[h!]
    \centering
    \begin{tabular}{lAAAAAA|AAAA}
    \toprule
        \multirow{2}{*}{} &  \multicolumn{2}{c}{VQP-VQ1} & \multicolumn{2}{c}{VQP-VQ2} & \multicolumn{2}{c}{VQP-VQ3} & \multicolumn{2}{c}{VQP-PE} & \multicolumn{2}{c}{VQP}\\
        & AUC & MDL & AUC & MDL & AUC & MDL & AUC & MDL & AUC & MDL\\
        \midrule
        \textbf{Pitch} & & & & & & & & & & \\
        Pitch & 0.588 & 63.01 & 0.801 & \textbf{50.04} & 0.586 & 63.07 & 0.870 & 42.13 & 0.742 & 55.78\\
        \midrule
        \textbf{Rhythm} & & & & & & & & & & \\
        Intensity & 0.682 & \textbf{59.15} & 0.681 & 59.58 & 0.640 & 61.49 & 0.797 & 50.01 & 0.662 & 60.97\\
        Num. sylls & 0.694 & \textbf{60.23} & 0.591 & 64.29 & 0.587 & 64.15 & 0.805 & 51.29 & 0.616 & 63.13\\
        \midrule
        \textbf{Tempo} & & & & & & & & & & \\
        Artic. rate  & 0.551 & \textbf{65.00} & 0.522 & 65.25 & 0.536 & 65.17 & 0.664 & 62.19 & 0.537 & 65.12\\
        Speech rate &  0.555 & 64.71 & 0.553 & \textbf{64.48} & 0.560 & 64.57 & 0.696 & 60.20 & 0.541 & 64.88\\
        Syll duration & 0.541 & \textbf{65.29} & 0.530 & 65.49 & 0.529 & 65.53 & 0.678 & 61.32 & 0.497 & 65.47\\
        Word duration & 0.805 & \textbf{51.36} & 0.735 & 57.35 & 0.690 & 59.41 & 0.988 & 14.55 & 0.749 & 54.58\\
        \midrule
        \textbf{Timbre} & & & & & & & & & & \\
        Formant f1 & 0.537 & 65.86 & 0.555 & \textbf{65.65} & 0.537 & 65.96 & 0.609 & 64.24 & 0.574 & 65.58\\
        Formant f2  & 0.516 & 65.64 & 0.523 & \textbf{65.62} & 0.516 & 65.65 & 0.547 & 65.28 & 0.514 & 65.60\\
        Formant f3  & 0.513 & 65.71 & 0.510 & \textbf{65.70} & 0.512 & 65.66 & 0.529 & 65.44 & 0.509 & 65.71\\
    \bottomrule
    \end{tabular}
    \caption{The disentanglement results from probing the VQP representations. We probe each of the three vector quantizer outputs (VQP-VQ1, VQP-VQ2 and VQP-VQ3) independently as per the setup in Table \ref{tab:explainability}, and recapitulate the VQP results for ease of comparison. We also show the probing results on the full output from the prosody encoder (VQP-PE).}
    \label{tab:disentanglement}
\end{table*}

Table \ref{tab:explainability} summarises the AUCs and minimum description lengths (MDLs) of probes, trained to predict subcomponents of prosody from different representations in the form of audio features (feature distributions are given in the supplementary materials). Whereas the baselines primarily encode timbral information, VQP selectively learns non-timbral prosody. This provides support for the hypotheses that underlie our inductive biases. Both primarily timbral and non-timbral prosodic representations obtain comparable performance on DAMMP, highlighting that there are multiple viable strategies for solving spoken language understanding tasks. VQP encodes complementary, new information compared with other speech representations.

We finally compare the representations in each part of the product quantizer, which is shown in Table \ref{tab:disentanglement}. Probes trained on contextualized VQP representations showed a similar pattern to the non-contextualized representations but generally have larger MDLs. We hypothesize this is because the ``raw'' prosodic information becomes more abstracted after the Transformer. The representations differed in how well the different prosodic features could be predicted from them, and importantly the features clustered based on prosodic domain for the different representations. Without using bottlenecks, the subcomponents of prosody naturally disentangle. Rhythm and tempo features were best predicted from VQP-VQ1, while pitch was best predicted from representation VQP-VQ2, corresponding to a dissociation of time and frequency features. Median word intensity was predicted similarly across the representations. Except for word duration, tempo features were not well represented by either factor representation, but notably they were still well represented by the combined representation from the product quantizer, demonstrating that encoding of prosody happens at different levels in the model, especially for these `derivative' features that require a combination of syllable and duration information. VQP-VQ3 was not immediately interpretable. It might be capturing something that is not represented in the features or act in a supporting function for the other representations. Despite downsampling the input, timbral features could be predicted to some degree, which could be due to an association with pitch not removed by the pitch correction or the $500\,$Hz limit in downsampling still capturing some spectral information; this is supported by the progressively worsening of performance with increasing formant number and that timbral features are predicted by VQP-VQ2, which also represents pitch the best.

\section{Conclusions}\label{sec:conclusions}
In this work, we introduced a self-supervised contrastive model that learns to selectively represent non-timbral prosody from raw audio without using bottlenecks, differing from prior subtractive approaches. Our approach retains competitive performance while making the prosody representations more de-identified than prior audio representations, which we quantify using an information-theoretic approach. We make explainability-related contributions by adopting probes for audio representations and demonstrating that product quantization can be used for disentanglement of prosody without using bottlenecks. Our work is motivated by building better real-world systems for spoken language understanding, where sufficient de-identifiability is crucial for user privacy and complying with HIPAA/GDPR regulation. One natural extension of our work is its application to building audio-linguistic representations.


\bibliography{references}
\bibliographystyle{include/icml2021}


\end{document}